%% file: ms.tex
\pdfoutput=1

\documentclass[a4paper,conference]{ieeeconf} 
\IEEEoverridecommandlockouts                             
\usepackage{fainekos-macros}
\usepackage{tabularx}
\usepackage{xfrac}
\usepackage{stmaryrd}
\usepackage[normalem]{ulem} 

\usepackage{url}
\usepackage{subcaption}

\newcommand\dptaliro{{\textsc{DP-TaLiRo}}\xspace}

\title{\LARGE \bf
Search-based Test-Case Generation by Monitoring \\ Responsibility Safety Rules
}

\author{Mohammad Hekmatnejad$^{1}$, Bardh Hoxha$^{2}$ and Georgios Fainekos$^{1}$
\thanks{This work has been partially funded by NSF awards 1350420 and 1361926, and the NSF I/UCRC Center for Embedded Systems.}
\thanks{$^{1}$Mohammad Hekmatnejad and Georgios Fainekos are with School of Computing, Informatics and Decision Systems Engineering, Arizona State University, Tempe AZ 85281, USA {\tt\small \{mhekmatn,fainekos\}@asu.edu }}%
\thanks{$^{2}$Bardh Hoxha is with Toyota Research Institute of North America, Ann Arbor, MI, 48187, USA
        {\tt\small bardh.hoxha@toyota.com}}%
}

\begin{document}

\maketitle
\thispagestyle{empty}
\pagestyle{empty}

\begin{abstract}
The safety of Automated Vehicles (AV) as Cyber-Physical Systems (CPS) depends on the safety of their consisting modules (software and hardware) and their rigorous integration. 
Deep Learning is one of the dominant techniques used for perception, prediction and decision making in AVs.
The accuracy of predictions and decision-making is highly dependant on the tests used for training their underlying deep-learning.  
In this work, we propose a method for screening and classifying simulation-based driving test data to be used for training and testing controllers.
Our method is based on monitoring and falsification techniques, which lead to a systematic automated procedure for generating and selecting qualified test data.
We used Responsibility Sensitive Safety (RSS) rules as our qualifier specifications to filter out the random tests that do not satisfy the RSS assumptions.
Therefore, the remaining tests cover driving scenarios that the controlled vehicle does not respond safely to its environment. 
Our framework is distributed with the publicly available \staliro and Sim-ATAV tools.
\end{abstract}

\input{introduction}

\input{preliminaries}

\input{rss}

\input{case-study}

\input{conclusion}

\bibliographystyle{IEEEtran}
\bibliography{mybib,fainekos_bibref,bh}

\end{document}

%% file: introduction.tex
\section{Introduction}


The development of safe, fully autonomous systems has become one of the seminal technological problems of our time.
Their eventual development will have a significant impact on mobility, with solutions ranging from personal mobility, to mobility-on-demand systems. 
The potential of automation carries a plethora of environmental and financial benefits \cite{nhtsa_ads}. 

Major components of these systems are developed to tackle problems of perception, decision making, and control, often using machine learning techniques such as Deep Neural Networks \cite{SiamEtAl2017itsc}. 
Ensuring that these components are bug-free is a challenging problem. 
To make the problem more difficult, emergent behavior that arises as a result of the complex interaction between these components is extremely difficult to detect
\cite{Lee2018ArsTechnica,JohnsonEtAl2015esv}. 


Physical testing has shown to be expensive, time consuming, and  generally inefficient.
In addition, there is no consensus on what constitutes an adequate metric in this area.
It is questionable whether metrics such as miles driven and miles per disengagement, which are typically reported to governmental agencies \cite{milesPerDiss2019}, are sufficient since most of the testing and miles driven are conducted in carefully chosen locations and scenarios. 
As a result, these metrics fail to capture a wide array of conditions and scenarios that are typical in day-to-day passenger driving.
 
Recent research \cite{TuncaliPF16itsc,CorsoEtAl2019itsc,dreossi2019verifai,majumdar2019paracosm,Abbas17Cyphy,KoschiEtAl2019itsc} supports simulation-based and virtual methods as a scalable alternative.
Simulation enables inexpensive, quick testing for arbitrarily complex conditions and scenarios. 
It also speeds up development by enabling automakers to quickly iterate and support the rapid development of their systems. 
In addition, since simulation-based methods have been studied extensively in the testing and verification community (see \cite{KapinskiEtAl2016csm} for an overview), it is only natural to attempt an extension to the autonomous vehicle domain. 


In \cite{fremont2019scenic}, the authors develop a probabilistic programming language \textsc{Scenic} to define scenes, which are configurations of physical objects in traffic environments.
The probabilistic component enables the assignment of behavior distributions to specific objects in the scene. 
In \cite{tuncali2019requirements,TuncaliEtAl2018iv}, the authors present Sim-ATAV, part of the \staliro{} \cite{fainekos2019robustness} toolbox, for adversarial testing of autonomous vehicles with machine learning components.
The tool enables the generation of test cases that falsify formal requirements expressed in Signal Temporal Logic (STL) \cite{BartocciEtAl2018survey}. 
Frameworks like \cite{dreossi2019verifai} and \cite{AbbasKRM2017asme} and the corresponding toolchains provide similar functionality to Sim-ATAV for search-based test generation.
A challenge encountered by all these methods is how to properly define what constitutes a violation of safe driving behaviors.

In this paper, Responsibility Sensitive Safety (RSS) \cite{shalev2017formal,GassmannEtAl2019iv,KoopmanOW2019safecomp} rules are utilized for classifying and qualifying driving test data to determine ego-centric and meaningful driving scenarios. In order to achieve that, RSS specifications are  formalized into STL to enable formal, algorithmic reasoning over them. 
In \cite{hekmat2019memocode}, RSS specifications were utilized to monitor naturalistic traffic data obtained from the CommonRoad library \cite{althoff2017commonroad}.
In this paper, the driving scenarios are generated through simulation. 
Given formal RSS specifications, a requirements-guided search through a stochastic optimization algorithm finds worst infractions of RSS specifications. 
The set of falsifying behaviors may be analyzed further by considering which part of the specification caused the falsification. 
This is beneficial since the root cause of failure may be determined and classified quickly. 
For example, using this approach, practitioners can return only the falsifying behaviors related to longitudinal distances between the ego car and other agents. 
To the best of the authors knowledge, this is the first time RSS specifications are formalized for testing vehicles in a simulated environment.

%% file: preliminaries.tex
\section{preliminaries}

\subsection{Signal Temporal Logic}

Signal Temporal Logic (STL) \cite{BartocciEtAl2018survey} was defined to express bounded time requirements over continuous-time (CT) signals. 
In this paper, we present the syntax and semantics of STL over discrete-time (DT) signals as resulting from simulations. 

\begin{defn}[STL Syntax for DT signals]
Let $x$ be a vector variable, i.e., $x = [x_1, \ldots, x_n]^T$, $p(x)$ be a function over the reals,
and $\mathcal{I}$ be any non-empty interval of  $\;\mathbb{R}_{\ge 0}$. 
The syntax for Signal Temporal Logic (STL) formulas is provided by the following grammar:
\begin{eqnarray*}
\phi ::= \Tb \;|\; p(x) \geq 0 \;|\; \neg \phi \;|\; \phi \vee \phi \;|\; \bigcirc_{\mathcal{I}} \phi \;|\; \phi \Un_{\mathcal{I}} \phi 
\end{eqnarray*}
where $\Tb$ is  {\it true}, $\bigcirc_{\mathcal{I}}$ is the next sample operator, and $\Un_{\mathcal{I}}$ is the until operator.
\end{defn}

We utilize a quantitative interpretation for the STL semantics (see \cite{BartocciEtAl2018survey} for an overview). 
In order to define quantitative semantics with a topological interpretation over arbitrary predicates $p(x)\geq 0$, we will need to use a \emph{generalized quasi-metric} $\genMet$ \cite{AbbasFSIG13tecs} (referred to simply as a metric for brevity in the following) to define a signed distance function:
\begin{defn}[Signed Distance] Let $x \in X$ be a point, $S \subseteq X$ be a set and $\genMet$ be a metric. 
Then, we define the Signed Distance from $x$ to $S$ to be 
	\[ \mathbf{Dist}_{\genMet}(x,S) := \left\{ \begin{array}{ll}
	- \inf\{\genMet(x,x')\;|\; x' \in S\} & \mbox{ if } x \not \in S \\
	\inf\{\genMet(x,x')\;|\; x' \not \in S\} & \mbox{ if } x \in S \\
	\end{array} \right. \]
\end{defn}
Intuitively, the distance function {returns} positive values when $x$ is in the set $S$ and negative values when $x$ is outside the set $S$. 
We should point out that we use the extended definition of supremum ($\sqcup$) and infimum ($\sqcap$).
That is to say, the supremum of the empty set is defined to be the bottom element of the domain, while the infimum of the empty set is defined to be the top element of the domain.
For example, when $\genVal = \CoRe_{\geq 0}$, then $\inf \emptyset := +\infty$. 

We review STL semantics that map a formula $\varphi$ and a trace $\dsig$ to a value drawn from a lattice $\overline{\genVal} \triangleq \genVal \cup \{-v \; | \; v\in \genVal \}$.
In this work, even though we need an arbitrary lattice of truth values $\genVal$ to treat Boolean signals, in most of the examples, we assume that  $\genVal = \CoRe_{\geq 0}$ (with the usual negation (-) over the reals).
We denote the robust valuation of the formula $\varphi$ over the trace $\dsig$ at sample $i$ by $\dle \varphi  \dri_{\genMet}(\dsig,i)$. 
The semantics for a predicate $p(x) \geq 0$ evaluated at time $i$ over trace $\dsig$ is defined as the distance between $\dsig(i)$ and the set $ \dle p(x) \geq 0 \dri \triangleq \{ x \, | \, p(x)\geq 0 \}$.  
Intuitively, this distance represents how robustly the point $\dsig(i)$ lies within (or is outside) the set $\dle p(x) \geq 0 \dri$. 
If this distance is zero, then  the smallest perturbation of the point $\dsig(i)$  can affect the outcome of $\dsig(i) \in \dle p(x) \geq 0 \dri$.  
 

\begin{defn} [Discrete-Time Robust Semantics] 
\label{def:rob-sem}
Consider an extended generalized quasi-metric space $(\genOut, \genMet)$.
Let $\dsig : N \rightarrow \genOut$ be a trace, then the robust semantics of an STL formula $\varphi$ with respect to $\dsig$ at time sample $i$ is  defined as:
\begin{eqnarray*} 
\dle \Tb \dri_{\genMet} (\dsig,i) & := & \bigsqcup \genVal := \top  
\\
\dle p(x)\geq 0  \dri_{\genMet} (\dsig,i) & := &
\mathbf{Dist}_{\genMet}(\dsig(i), \dle p(x)\geq 0 \dri) 
\\
\dle \neg \varphi_1 \dri_{\genMet}  (\dsig,i) & := & - \dle \varphi_1 \dri_{\genMet}  (\dsig,i) 
\\
\dle \varphi_1 \vee \varphi_2 \dri_{\genMet}  (\dsig,i) & := & \dle \varphi_1 \dri_{\genMet}  (\dsig,i) \sqcup \dle \varphi_2 \dri_{\genMet}  (\dsig,i)
\end{eqnarray*}
\begin{align*}
\begin{split}
&\dle \bigcirc_{\mathcal{I}} \varphi_1 \dri_{\genMet}  (\dsig,i)  := 
\\
&\left \{
  \begin{tabular}{ll}
  $\dle \varphi_1 \dri_{\genMet}  (\dsig,i+1)$ & \; \footnotesize{if  $i+1 \in N \;and\; (i+1) \Delta t  \in (i \Delta t + \Ic)$}\\ 
  $-\top  \triangleq \bot$ & \; \footnotesize{otherwise (i.e., the bottom element)} \\
  \end{tabular}
\right .
\end{split}\\
\begin{split}
&\dle \varphi_1 \Un_\Ic \varphi_2 \dri_{\genMet}  (\dsig,i)  :=
 \\
&\bigsqcup_{i' \in \{ j \in N \; | \; j \Delta t \in (i\Delta t+\Ic) \} } \left ( \dle \varphi_2 \dri_{\genMet}  (\dsig,i') \sqcap  \bigsqcap_{i \leq i'' <i'} \dle \varphi_1 \dri_{\genMet}  (\dsig,i'')
\right )
\end{split}
\end{align*}
where $\top$ is the top element of the lattice, and $t+\Ic = \{ t'' \; | \; \exists t' \in \Ic \, .\, t'' = t + t' \}$.

\label{def:mitlrob}
\end{defn}

Intuitively, the requirement $\bigcirc_{\mathcal{I}} \varphi$ states that $\varphi$ should be true at the next sample, which should occur some time in the physical time interval $\Ic$. 
For example, consider $\Delta t =0.1$ and the formula $\psi = \bigcirc_{[0,0.1]} \mathbf{T}$, then $\psi$ is true ($\top$) at sample $i$ since $(i+1)\Delta t - i\Delta t = \Delta t \in [0,0.1]$.
However, for $\Delta t = 0.2$, $\psi$ would evaluate to false ($\bot$).
The operator $\varphi_1 \Un_{\mathcal{I}} \varphi_2$ states that $\varphi_2$ should be satisfied at some time in the interval $\Ic$ and until then $\varphi_1$ should hold.
The other common Boolean and temporal operators can be defined as syntactic abbreviations (see \cite{BartocciEtAl2018survey}).
For example, $\Diamond_\Ic \phi \equiv {\mathbf T} \Un_\Ic \phi$  stands for eventually at some time in the time interval ${\mathcal{I}}$, $\phi$ should be true, and  $\Box_\Ic \phi \equiv \neg \Diamond_\Ic \neg \phi$ stands for always during the interval ${\mathcal{I}}$,  $\phi$ should be true.
When ${\mathcal{I}} = [0,\infty)$, we will be dropping ${\mathcal{I}}$ from the notation, e.g., $\Box_{[0,\infty)} \phi \equiv \Box \phi$.

An important operator that we will be using in this work is the {\it release} operator $\varphi_1 \Rc_{\mathcal{I}} \varphi_2$ $\equiv $ $\neg ( \neg \varphi_1 \Un_{\mathcal{I}} \neg \varphi_2)$, which states that $\varphi_2$ should always hold during the time interval $\mathcal{I}$ up to (but not including) the time when $\varphi_1$ becomes true.
In fact, we will need a slightly modified version of the release operator $\rRl$ which does not require $\varphi_2$ to happen at all if $\varphi_1$ has happened in the past:
\begin{multline*}
\dle \varphi_1 \rRl_\Ic \varphi_2 \dri_{\genMet}  (\dsig,i)  := 
\\
\bigsqcap_{i' \in \{ j \in N \; | \; j \Delta t \in (i\Delta t+\Ic) \} } \left ( 
\dle \varphi_2 \dri_{\genMet}  (\dsig,i') \sqcup  \bigsqcup_{i \leq i'' \leq i'} \dle \varphi_1 \dri_{\genMet}  (\dsig,i'')
\right )
\end{multline*}
We refer to the above operator as non-strict release operator. In fact, any non-strict release formula such as $\varphi_1 \rRl_\Ic \varphi_2$ can be rewritten as $\varphi_1 \Rc_\Ic (\varphi_1 \vee \varphi_2)$ using the release operator.

\subsection{Robustness-guided Falsification of STL Specifications}

The problem of determining whether a CPS $\Sigma$ satisfies a specification $\phi$ is an undecidable problem, i.e. there is no general algorithm that terminates and returns whether $\Sigma \models \phi$. Therefore, it is not possible to determine exactly the minimum robustness over all system behaviors. However, by repeatedly testing the system, we can check whether a behavior that does not satisfy the specification exists. 

Falsification is the process of finding a counter-example or falsifying example that proves that the system does not satisfy the specification. The STL falsification problem is presented as follows:

\begin{prob}[STL Falsification]
Given an MTL formula $\phi$ and a system $\Sys$, find initial conditions and input signals such that, when given to $\Sys$, generate a falsifying trajectory for which $\dle \phi \dri (\tss) <0 $ (or with Boolean semantics $\tss \not\models \phi$).\end{prob}

The robustness semantics of STL enable the development of an optimization framework that searches for falsifying behavior in a system. Specifically, the problem may be posed as a global non-linear optimization problem:

\begin{eqnarray} \label{eqn:T_star}
\simTr^\star = \arg \min_{\simTr \in \mathcal{L}(\Sigma)}  \robsemD{\varphi}(\simTr).
\end{eqnarray}

Here, \simTr is a system trajectory and $\mathcal{L}(\Sigma)$ is the set of all system behaviors. 
If the global optimization algorithm converges to some local minimize $\tilde \simTr$ such that $\robsemD{\varphi}(\tilde \simTr)<0$, then a counterexample (adversarial sample) has been identified, which can be used for debugging (or for training).

In order to solve this non-linear non-convex optimization problem, a number of stochastic search optimization methods can be applied (e.g., \cite{AbbasFSIG13tecs} -- for an overview see \cite{KapinskiEtAl2016csm}).
We leverage existing falsification methods to identify falsifying examples for an autonomous driving system.

%% file: rss.tex
\section{RSS}

In this paper, our focus is to use the RSS specifications for monitoring the safety of driving behavior of an ego car \cite{shalev2017formal}.
In our test environment, non-ego cars' behaviors are as unconstrained as possible. 
Hence, worst-case scenarios are not always meaningful.
We utilize the RSS requirements to place constraints solely on the ego car's behavior and not other vehicles in the road (referred to as agent vehicles from now on). 

Before diving into a formal definition, we first informally describe the RSS requirements for an ego car driving on the highway, and then formally represent them in STL.
A situation is unsafe if and only if the longitudinal and lateral distances among the ego vehicle and agent vehicles are unsafe.
By lateral and longitudinal safe distances, we refer to the minimum required distances, which are calculated based on the current velocities of the ego car, ego’s maximum reaction time ($\rho$), and the maximum/minimum allowed accelerations stated in Table \ref{tbl:params}.

\begin{figure}[h]
  \centering
  \includegraphics[width=1\linewidth]{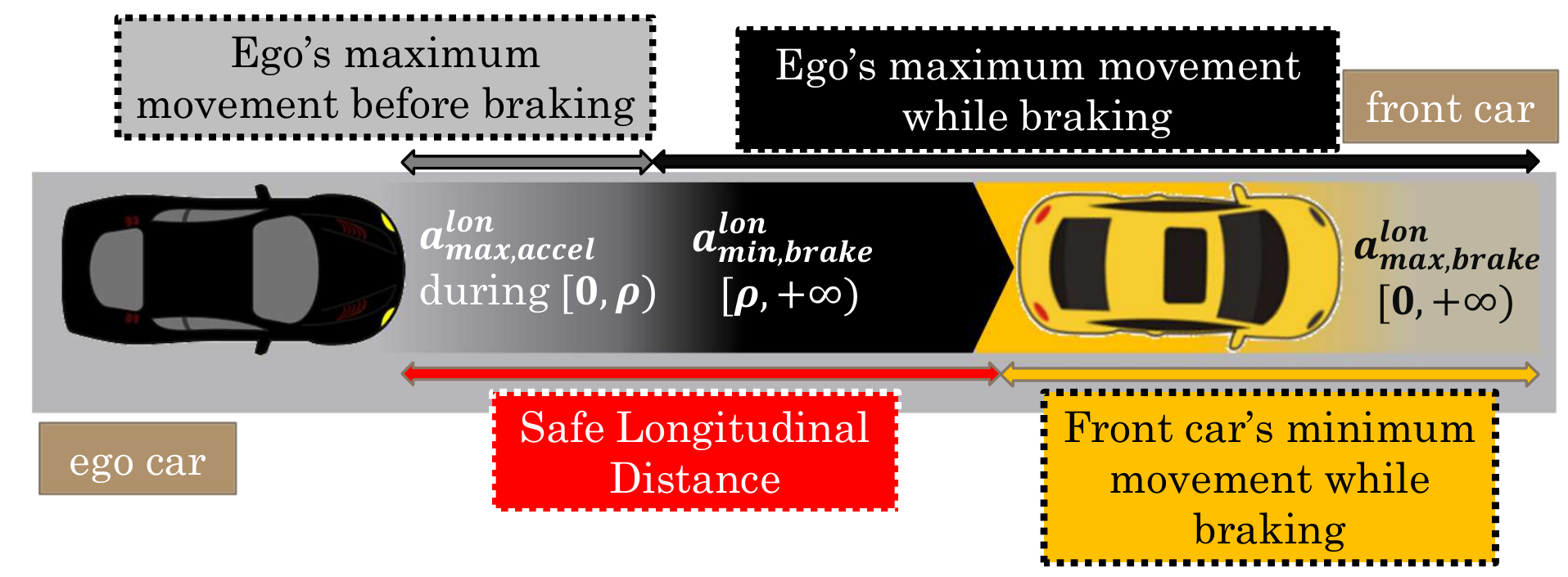}
  \caption{{Longitudinal Safety. See Table \ref{tbl:rss-param-desc} for the symbol definitions.}}\label{fig:long-safety}
\end{figure}

\begin{figure}[h]
  \centering
  \includegraphics[width=1\linewidth]{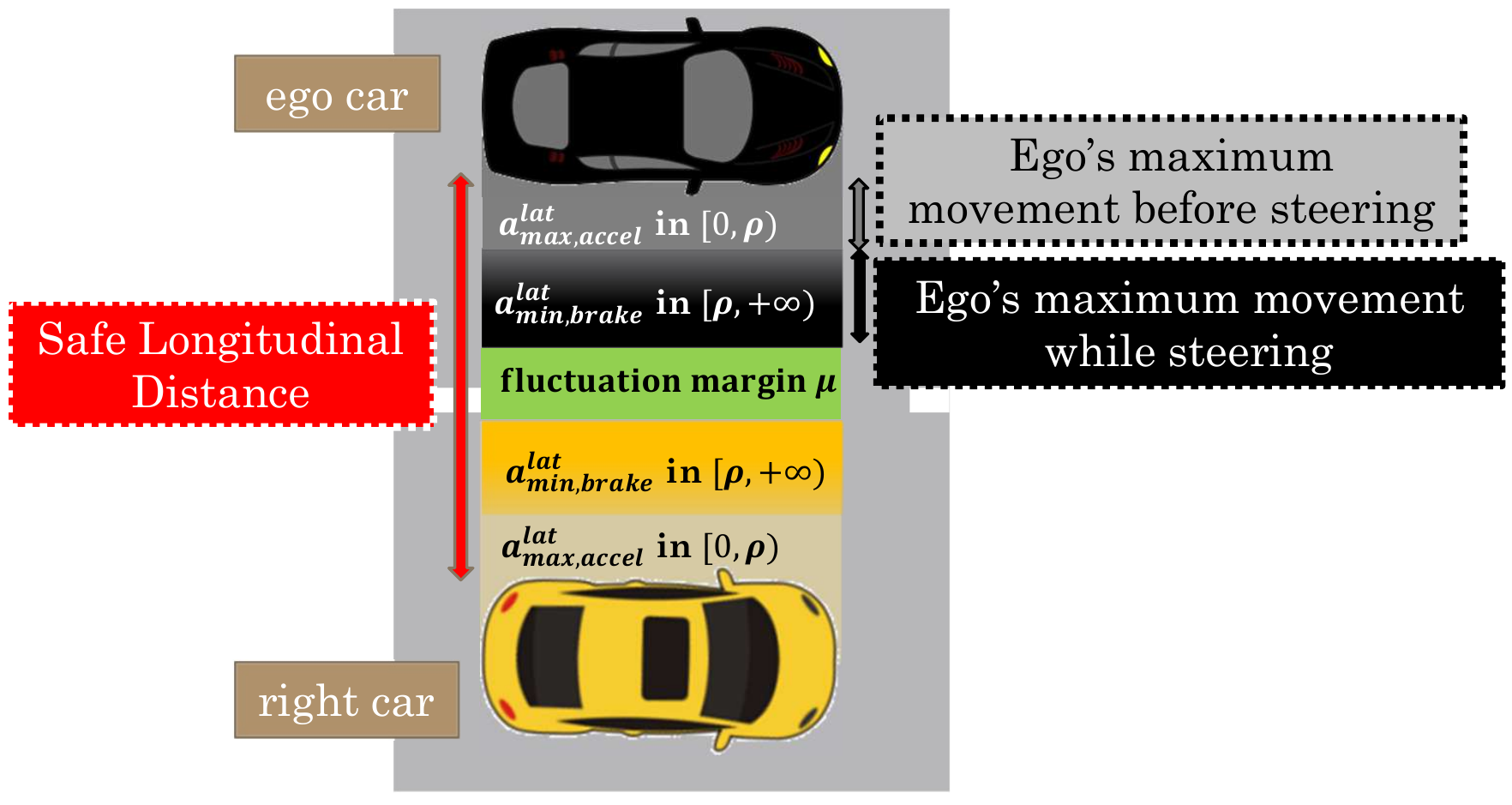}
  \caption{{Lateral Safety. See Table \ref{tbl:rss-param-desc} for the symbol definitions.}}\label{fig:lat-safety}
\end{figure}

We illustrate the concept of safe longitudinal distance and predicates needed in our definitions in Fig. \ref{fig:long-safety}.
Similarly, in Fig. \ref{fig:lat-safety}, we illustrate the concept of safe lateral distance and related predicates.
The idea is that if an ego car continuously preserves the minimum required safe distances with a leading agent vehicle until the distances become unsafe, then it can react responsibly and responsively in order to reach a safe state again.
There are two significant restrictions on the ego’s driving behavior based on time.
We use \textit{hesitation} and \textit{reaction} times to refer to times that fall in $[0,\rho)$ and $[\rho,+ \infty]$, respectively.
For all the hesitation times, the ego car cannot exceed a maximum lateral/longitudinal acceleration bound depending on which distance changed from safe to unsafe.
For all reaction times, the ego car has to react responsibly. 
That is, the ego car has to start decelerating with the minimum required deceleration depending on which distance changed from safe to unsafe.
For the case that a situation becomes unsafe laterally,  there is another restriction on the velocity of the ego car.
That is, in the reaction time, the ego car has to control its lateral velocity so that it never deviates more than a constant $\mu/2$ from its original lateral position when the situation became unsafe laterally. 
All the restrictions are lifted when an unsafe situation becomes safe again.
For more details see \cite{shalev2017formal} and \cite{hekmat2019memocode}.

\begin{table}
\begin{center}
\caption{{
The RSS parameter description in the safe lateral and longitudinal definitions.
}}\label{tbl:rss-param-desc}
\resizebox{.49\textwidth}{!}{
\begin{tabular}{ |c|c| } 
\hline
\textbf{symbol} & \textbf{description} \\
\hline
$\rho$ & reaction time in $seconds$ \\ [0.1cm]
$\mu$ & minimum lateral distance in $meter$\\[0.1cm]
$\delta_t$ & sampling time in $seconds$ \\ [0.1cm]
$a_{minBr}^{lon}$ & minimum required longitudinal braking in $m/s^2$\\[0.1cm]
$a_{maxAcc}^{lon}$ & maximum allowed longitudinal accelerating in $m/s^2$ \\ [0.1cm]
$a_{minBr}^{lat}$ & minimum required lateral braking in $m/s^2$\\ [0.1cm]
$a_{maxBr}^{lon}$ & maximum allowed longitudinal braking in $m/s^2$ \\ [0.1cm]
$a_{maxAcc}^{lat}$ & maximum allowed lateral accelerating in $m/s^2$\\ [0.1cm]
\hline
\end{tabular}
}
\end{center}
\end{table}

In the rest of this section, we represent a slightly modified version of the formula $\varphi_{resp}^{lat,lon}$ from the Remark  III.3 in \cite{hekmat2019memocode}.
For the sake of space and readability, for the lateral requirements, we assumed that the ego car is driving on the left side of the other car.
Also, we assume that the car behind the front car is the ego car.
\begin{defn}[RSS formalization in STL for a single ego car driving in highway along other cars]\label{def:rss}

The below formula $\varphi^{lat,lon}_{resp}$ enforces the required RSS restrictions on the velocity and accelerations of an ego car driving on a highway.

\begin{gather*}
\varphi^{lat,lon}_{resp} \equiv \varphi^{lon} \wedge \varphi^{lat} \wedge \varphi^{lat,lon} \wedge \varphi^{\neg lat,\neg lon}\\
\varphi^{lon} \equiv  \\
\Box 
\biggl ( 
\bigl(\neg S_{l,r}^{lat} \wedge S_{b,f}^{lon} \wedge \bigcirc (\neg{S_{l,r}^{lat}} \wedge \neg{S_{b,f}^{lon}})\bigr )
\rightarrow 
\bigcirc
 P^{lon} \biggr)\\
 \varphi^{lat} \equiv \\
\Box 
\biggl( 
\bigl(S_{l,r}^{lat} \wedge \neg S_{b,f}^{lon} \wedge \bigcirc (\neg{S_{l,r}^{lat}} \wedge \neg{S_{b,f}^{lon}})\bigr )
\rightarrow \bigcirc P^{lat}
\biggr )\\
\varphi^{lat,lon} \equiv\\ 
\Box\biggl( \bigl(S_{l,r}^{lat} \wedge S_{b,f}^{lon} \wedge \bigcirc (\neg{S_{l,r}^{lat}} \wedge \neg{S_{b,f}^{lon}}) \bigr)
\rightarrow \\
\bigcirc (P^{lat} \vee  P^{lon})\biggr)\\
\varphi^{\neg lat,\neg lon} \equiv\\ 
\biggl( 
(\neg{S_{l,r}^{lat}} \wedge \neg{S_{b,f}^{lon}}) 
\rightarrow
\bigcirc 
(P^{lat} \vee  P^{lon})\biggr)
\end{gather*}

where $P^{lon}$ and $P^{lat}$ are defined as 
\begin{gather*}
 P^{lon} \equiv 
 \bigl(
  (S_{l,r}^{lat} \vee S_{b,f}^{lon})
  \rRl_{[0,\rho)}( A_{b,maxAcc}^{lon} 
 ) 
 \wedge
\\
(S_{l,r}^{lat} \vee S_{b,f}^{lon})
\rRl_{[\rho,+\infty)} (A_{b,minBr}^{lon} 
\bigr)\\
P^{lat} \equiv \bigl( P^{lat}_{0,\rho} \wedge P^{lat,1}_{\rho,\infty} \wedge P^{lat,2}_{\rho,\infty}\bigr)     
\end{gather*}
and the subformulas $P^{lat}_{0,\rho}$, $P^{lat,1}_{\rho,\infty}$ and $P^{lat,2}_{\rho,\infty}$ are defined as 
\begin{gather*}
P^{lat}_{0,\rho} \equiv  
(S_{l,r}^{lat} \vee S_{b,f}^{lon})
\rRl_{[0,\rho)} ( A_{l,maxAcc}^{lat} 
)  \\
P^{lat,1}_{\rho,\infty} \equiv 
\bigl ( 
S_{l,r}^{lat} \vee S_{b,f}^{lon}
\vee V_{l,stop}^{lat}  
\bigr ) 
\rRl_{[\rho,+\infty)} A_{l,minBr}^{lat} 
\\
P^{lat,2}_{\rho,\infty} \equiv  
(S_{l,r}^{lat} \vee S_{b,f}^{lon})
\rRl_{[\rho,+\infty)} 
\bigl ( 
V_{l,stop}^{lat} 
\rightarrow \\
(S_{l,r}^{lat} \vee S_{b,f}^{lon})
\rRl (V_{l,neg}^{lat}) 
\bigr) 
\\
\end{gather*}
By $V_{l,neg}^{lat}$ we mean that the $\mu-$lateral velocity of the ego car is less than or equal to zero. 
\end{defn}

Note that all the predicates' definitions and assumptions can be found in \cite{hekmat2019memocode}.
The formula $\varphi_{resp}^{lat,lon}$ consists of four subformulas such as for each a unique trigger or response is defined.
The subformula $\varphi^{lon}$ gets triggered if a safe longitudinal and an unsafe lateral distances become unsafe all together. 
We call the time that both lateral and longitudinal distances become unsafe, a dangerous time.
If time $t_b$ is a new dangerous time and after that until the current time $t$ all times in $[t_b,t]$ are dangerous times, then we call $t_b$ the Dangerous Threshold time as in \cite{shalev2017formal}.
The desired response for this case enforces some longitudinal restrictions on the behavior of the ego car.
Similarly, the subformula $\varphi^{lat}$ gets triggered if an unsafe longitudinal and a safe lateral distances become unsafe all together.
The desired response for this case enforces some lateral restrictions on the behavior of the ego car.

Unlike the $\varphi^{lon}$ and $\varphi^{lat}$ subformulas, the next two subformulas $\varphi^{lat,lon}$ and $\varphi^{\neg lat,\neg lon}$ are designed to cover the cases that either 
a situation changes from completely safe to unsafe, or it starts completely unsafe.
The desired response to both cases requires either or both responses of the two previous responses.


%% file: case-study.tex
\section{RSS Application on Automated Test Generation and Classification}

By using the formalized RSS requirements as STL formulas, we can collect more meaningful samples for testing purposes. 
More specifically, we are interested in screening test driving scenarios and classifying them based on safety conditions introduced by RSS. 
One advantage of automated and guided test generation and classification is to only generate automatically annotated tests for testing and training perception based systems in ADS.

In this section, we show how one can utilize the formalized RSS requirements to select 
useful scenarios among a set of arbitrarily generated driving scenarios for testing purposes.
We used \staliro and Sim-ATAV in our experiments for monitoring and simulating trajectories, respectively \cite{taliro_tools,fainekos2019robustness,tuncali2019requirements}.

To show the effectiveness of our proposed framework, we represent and use a basic collision avoidance specification (CAS from now on) concerning the distances of an ego vehicle and other agent vehicles.
We compare the result of monitored scenarios by using the CAS and the RSS requirements separately.
We formalized the CAS requirements in STL representation in the following definition.

\begin{defn}[Collision Avoidance Specification]\label{def:cav}
We call two objects collided if and only if their shortest distance get closer than a $\delta$. 
The STL version of the requirement is shown as below:
\begin{gather*}
    \varphi_{cas} \equiv 
    \Box \neg 
    \bigl(
        dx_{ego}^{a1}<\delta_x \wedge 
        dy_{ego}^{a1}<\delta_y \wedge \\
        dx_{ego}^{a2}<\delta_x \wedge 
        dy_{ego}^{a2}<\delta_y
    \bigr)
\end{gather*}
where Table \ref{tbl:cas-param-desc} represents the symbol descriptions. 
\end{defn}

For the tests, we only considered highway traffic scenarios, which by design do not include intersections, and wherein all the self-driving vehicles are moving in the same direction. 
In the scenarios, there are two agent vehicles and one ego vehicle.
We generated our test scenarios with some restrictions for the initial positions, orientations, and speeds of the vehicles. 
Also, we enable the two agent vehicles to drive along the highway and maneuver randomly.
We used the information in Table \ref{tbl:params} as our test generating configuration parameters.

Note that our approach is not restricted to the presented scenarios, and it applies to the more general traffic scenarios which are discussed in \cite{shalev2017formal}. 
We used \dptaliro \cite{taliro_tools,fainekos2012verification} as our offline monitoring tool for computing the robustness of the traffic scenarios.

\begin{table}
\begin{center}
\caption{{
The CAS symbol description in the safe lateral and longitudinal definitions. All distances are in meter.
}}\label{tbl:cas-param-desc}
\resizebox{.5\textwidth}{!}{
\begin{tabular}{ |c|c| } 
\hline
\textbf{symbol} & \textbf{description} \\
\hline
$dx_{ego}^{a1}$ & longitudinal distance between the go vehicle and the adversarial vehicle 1 \\ [0.1cm]
$dx_{ego}^{a2}$ & longitudinal distance between the go vehicle and the adversarial vehicle 2 \\ [0.1cm]
$dy_{ego}^{a1}$ & lateral distance between the go vehicle and the adversarial vehicle 1 \\ [0.1cm]
$dy_{ego}^{a2}$ & lateral distance between the go vehicle and the adversarial vehicle 2 \\ [0.1cm]
$\delta_{x}$ & a constant safe longitudinal distance \\ [0.1cm]
$\delta_{y}$ & a constant safe lateral distance \\ [0.1cm]
\hline
\end{tabular}
}
\end{center}
\end{table}

\begin{table}
\begin{center}
\caption{{
The parameter values in the case-study and experiments.
}}\label{tbl:params}
\resizebox{.4\textwidth}{!}{
\begin{tabular}{ |c|c|c|c| } 
\hline
\textbf{parameter} & \textbf{value range} & \textbf{parameter} & \textbf{value range}\\
\hline
$x_{init}^{ego}$ & [0,5] $m$ & $y_{init}^{ego}$ & [-1.5,4.5] $m$\\[0.1cm]
$\theta_{init}^{ego}$ & [-$\pi/8$,$\pi/8$] & $v_{init}^{ego}$ & [10,25] $m/s$\\[0.1cm]
$x_{init}^{a1}$ & [20,40] $m$ & $y_{init}^{a1}$ & [-3.5,-1.5] $m$\\[0.1cm]
$x_{init}^{a2}$ & [10,25] $m$ & $y_{init}^{a2}$ & [1,4] $m$\\[0.1cm]
$v_{init}^{a1}$ & [5,15] $m/s$ & $v_{init}^{a2}$ & [0,30] $m/s$\\[0.1cm]
$y^{a1}$ & [-5,0] $m$ & $v^{a1}$ & [10,30] $m/s$\\[0.1cm]
$y^{a2}$ & [0,5] $m$ & $v^{a2}$ & [4,20] $m/s$\\[0.1cm]
\hline
\end{tabular}
}
\end{center}
\end{table}

\begin{table}
\begin{center}
\caption{{
The RSS parameter values in the case-study and experiments.
}}\label{tbl:rss-params}
\resizebox{.35\textwidth}{!}{
\begin{tabular}{ |c|c|c|c| } 
\hline
\textbf{parameter} & \textbf{value} & \textbf{parameter} & \textbf{value}\\
\hline
$\rho$ & 0.5 $sec$ & $\mu$ & 0.4 $m$\\[0.1cm]
$\delta_t$ & 0.01 $sec$ & $a_{minBr}^{lon}$ & 4 $m/s^2$\\[0.1cm]
$a_{maxAcc}^{lon}$ & 4.5 $m/s^2$ & $a_{minBr}^{lat}$ & 3 $m/s^2$\\ [0.1cm]
$a_{maxBr}^{lon}$ & 2.5 $m/s^2$ & $a_{maxAcc}^{lat}$ & 3 $m/s^2$\\ [0.1cm]
\hline
\end{tabular}
}
\end{center}
\end{table}

\subsection{Vehicle Controller}
In our experiments, we used an existing Sim-ATAV vehicle controller.
The controller receives (provided by simulator) camera, LIDAR, and Radar sensor information as to its input signals.
There is a perception system as part of the controller that performs sensor data processing and sensor fusion.
The controller is capable of detecting object types (vehicle, pedestrian) and their predicted trajectories.
The control algorithm sends signals to the throttle and steering actuators.
The controller's goal is to safely follow a predetermined path.
For more information, we refer readers to \cite{tuncali2019requirements}.
We chose $25m/s$ as the max and target velocity for the ego vehicle. 

\subsection{Simulated Tests and Environment}
In our experiments, the simulated environment consists of a straight highway with three lanes.
The initial test conditions concern position, orientation, and velocity of the ego vehicle.
The initial test conditions for the agent vehicles concern the same parameters as the ego vehicle, except that their orientations are fixed.

\begin{figure}[h]
  \centering
  \includegraphics[width=0.9\linewidth]{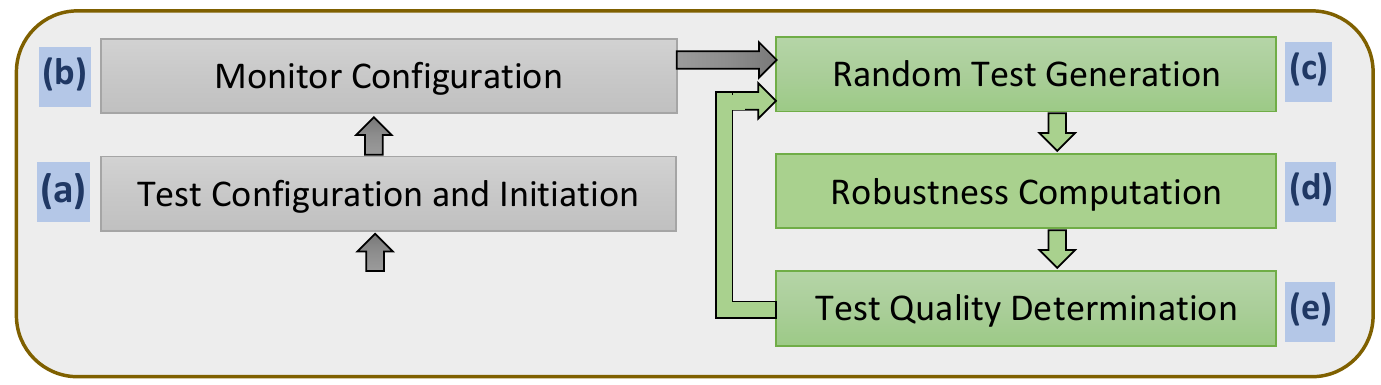}
  \caption{{The flowchart represents the executing order of major tasks in our framework.}}\label{fig:framework}
\end{figure}

At each run of test generation, a uniform random sampler randomly chooses the initial conditions of all the vehicles.
It also generates random trajectories for agent vehicles.
In this experiment, we predetermined a fixed trajectory for the ego vehicle.
However, the ego vehicle’s controller is free to do collision avoidance maneuvers.
We ran a (hit-and-run) sampler to produce and store one thousand of sample scenarios for the test parameters shown in Table \ref{tbl:params}.
We chose the parameter values for demonstration purposes and they might not reflect realistic values.
During the rest of the section, we refer to the code-base that controls test screening and classifying \textit{test control unit}.
The details of tasks in Fig. \ref{fig:framework} are explained in the following:
\begin{itemize}
    \item \textit{(a)}: 
    The very first step is to design test scenarios and assign range values for configurable parameters. 
    For example, we used data in the Table \ref{tbl:params} in our experiments. 
    Also, we used Sim-ATAV to design driving scenarios.
    \item \textit{(b)}: 
    We use \staliro as our monitoring tool, and configure it based on the our test design. 
    Two important tasks here are to formalize a desirable specification in STL form and define input signals. 
    Here we use two STL formulas in Def. \ref{def:rss} and Def. \ref{def:cav} for comparison purposes. 
    \item \textit{(c)}: 
    The \staliro tool is equipped with random samplers and optimizers. 
    We use the random-walk algorithm from the \staliro to generate uniformly random test data. 
    When optimization is needed, we use Simulated Annealing with hit and run Monte Carlo sampling.
    The test data consists of all the initial values for the vehicles (as in Table \ref{tbl:params}) that are needed to run the designed simulation.
    The test control unit passes the generated test to the Sim-ATAV through an interface.
    Sim-ATAV initializes the simulator (Webots) and makes it ready to run the test.  
    When the simulation terminates, the Sim-ATAV passes the simulated trajectories of all the vehicles to the test control unit.
    Next, the test control unit translates the trajectories into the lane-based coordinate system that the RSS rules are defined based on.
    Lastly, using the translated trajectories, we compute the final input signals to the monitor. 
    That is, each signal represents a corresponding predicate in the RSS formula such as safe distances or $\mu$-lateral velocities.
    \item \textit{(d)}: 
    Our \dptaliro monitor computes the robustness of the generated driving trajectories (input signal) for the ego vehicle given an STL specification formula.
    The positive or negative sign of the computed robustness value states if the specification is satisfiable or falsifiable, respectively. 
    The magnitude of the robustness value determines the resistance of the signal against perturbations. 
    \item \textit{(e)}:
    In this experiment, we are interested in accumulating test scenarios that 
    do not meet safety standards based on RSS rules.
    If the robustness of the RSS specification was negative, then the test driving scenario does not satisfy the RSS requirements, and, therefore, it is a useful sample.
\end{itemize}


\subsection{Case Study}
\begin{figure}[tbp]
\vspace{6pt}
\centering
\begin{subfigure}{1\linewidth}
  \centering
  \includegraphics[width=1\linewidth]{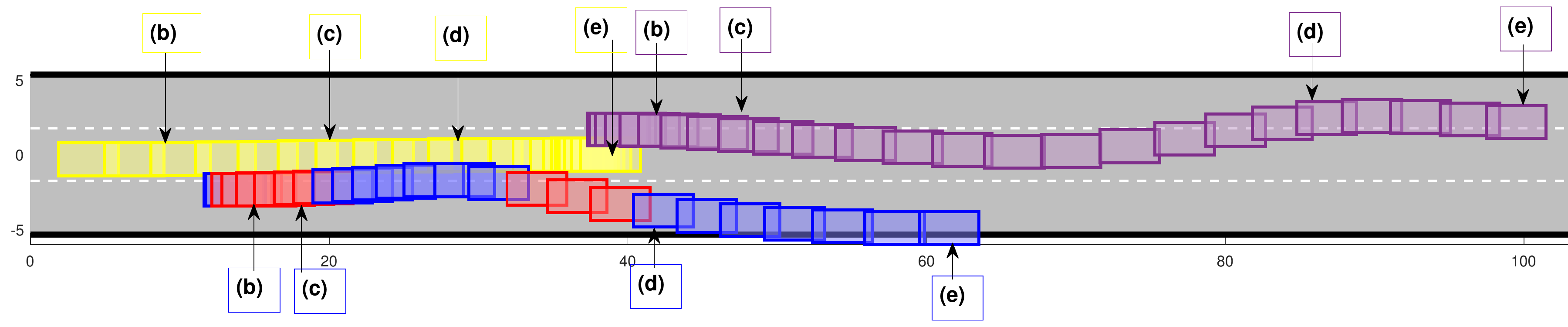}
  \caption{\scriptsize{A sample trajectory of three vehicles. The ego vehicle's trajectory is shown in the middle lane and colored yellow. The red part of the vehicle's trajectory in the lowest lane highlights the close distance to the ego vehicle. The (b),(c),(d), and (e) refer to the times that below snapshots are taken.}}\label{fig:sample-traj}
\end{subfigure} 
\begin{subfigure}{.45\linewidth}
  \centering
  \includegraphics[width=1\linewidth]{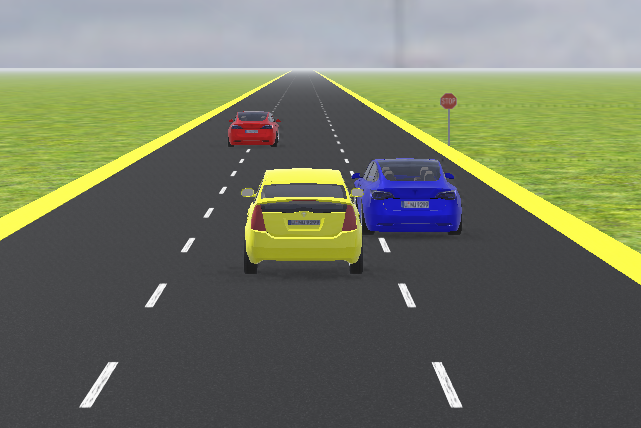}
  \caption{\scriptsize{first snapshot.}}\label{fig:scn-1}
\end{subfigure} 
\begin{subfigure}{.45\linewidth}
  \centering
  \includegraphics[width=1\linewidth]{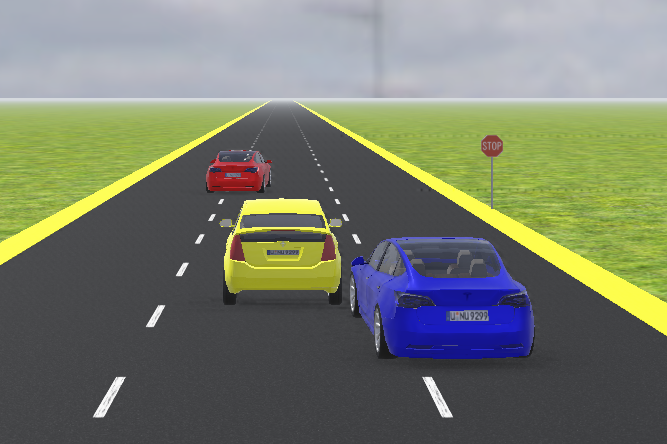}
  \caption{\scriptsize{second snapshot.}
  }\label{fig:scn-2}
\end{subfigure} 
\begin{subfigure}{.45\linewidth}
  \centering
  \includegraphics[width=1\linewidth]{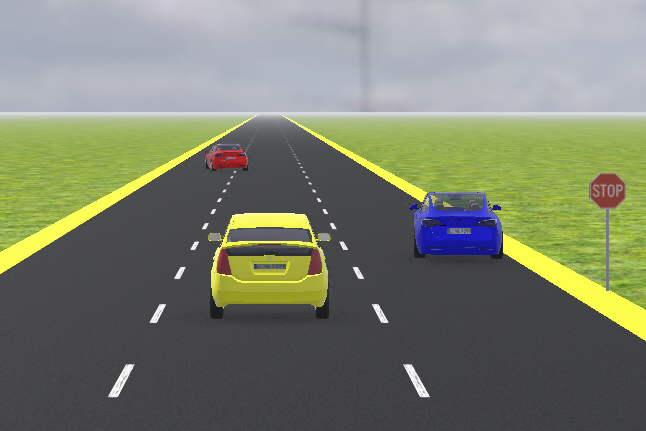}
  \caption{\scriptsize{third snapshot.}}\label{fig:scn-3}
\end{subfigure} 
\begin{subfigure}{.45\linewidth}
  \centering
  \includegraphics[width=1\linewidth]{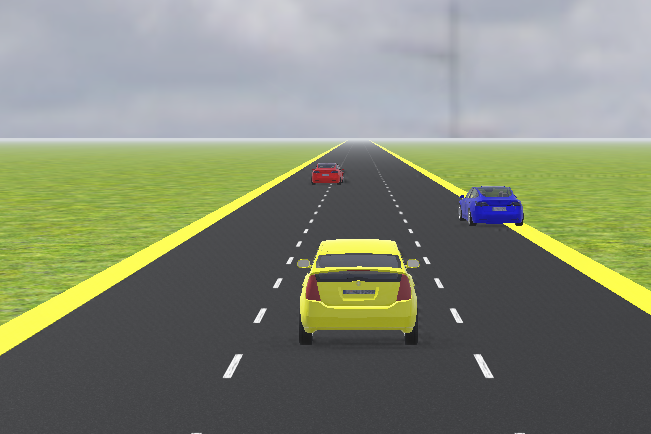}
  \caption{\scriptsize{forth snapshot.}
  }\label{fig:scn-4}
\end{subfigure} 
\caption{A sample traffic scenario taken from Webots.}
\label{fig:case-study}
\end{figure}

We chose some interesting scenarios to explain and analyze here.
In these scenarios we show how the RSS specification helps to select useful tests for verifying the vehicle's controller. 
The first scenario was test number 131 among 1000 uniformly generated tests.
Figure \ref{fig:sample-traj} represents half of the trajectory in the test (from time $t=0$ to time $t=5$).
The colors of boxes in each trace represents a vehicle in the simulation.
The ego vehicle with yellow color drives in the middle lane.
The two other vehicles in magenta and blue drive to the left and right of the ego vehicle.
We highlighted the part of the trajectories that were close to the ego vehicle with red color.
Also, under the trajectory diagram in Fig. \ref{fig:case-study}, there are four snapshots taken from the simulation at different times.

We monitored the specification in Def. \ref{def:rss} for the above scenario. 
The robustness result was negative, and the monitor reported the $a^{lon}_{maxAcc}$ in subformula $P^{lon}$ as the predicate that caused the maximum negative robustness value.
This was due to high acceleration by the ego vehicle in the beginning of the scenario.
The ego vehicle starts braking as soon as its controller detects the leading vehicle and assigns a high level of caution to it. 

However, during the hesitation period, it did exceed the required maximum acceleration.
Interestingly, this case did not violate the RSS rules for the blue vehicle driving closely 
to the right, but because of the red vehicle on the left of it.

\begin{figure}[tbp]
\vspace{6pt}
\centering

\begin{subfigure}{.45\linewidth}
  \centering
  \includegraphics[width=1\linewidth]{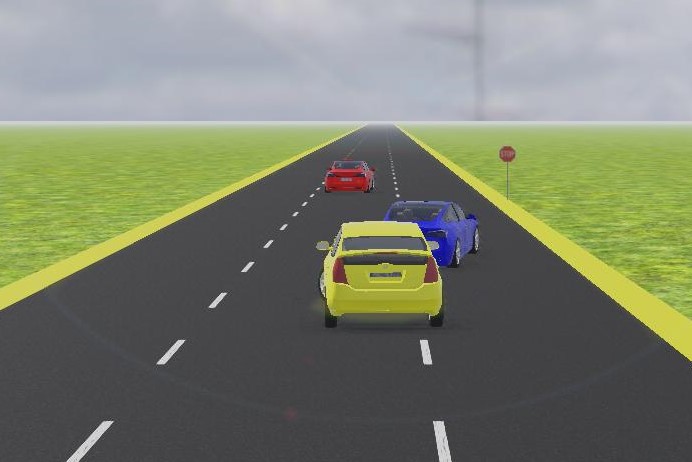}
  \caption{\scriptsize{A scenario starts in an unsafe longitudinal and safe lateral situation, and then it also becomes unsafe laterally when the cars change their lane.}}\label{fig:scn-dangerous}
\end{subfigure} 
\begin{subfigure}{.45\linewidth}
  \centering
  \includegraphics[width=1\linewidth]{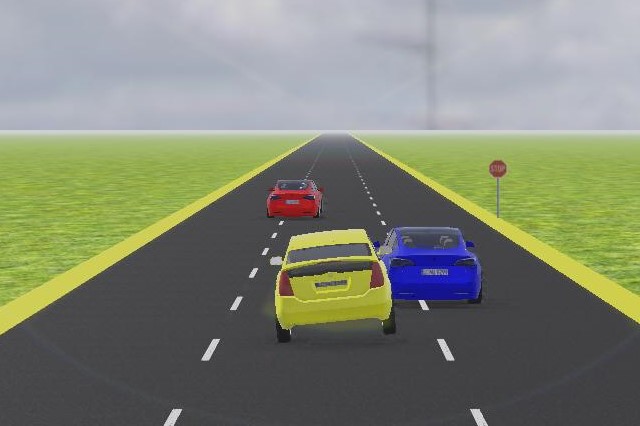}
  \caption{\scriptsize{A scenario starts unsafe laterally and longitudinally. 
  The blue agent vehicle turns in front of the ego vehicle and causes a virtual collision.
  }}\label{fig:scn-accident}
\end{subfigure} 
\caption{Two common unsafe scenarios.}
\label{fig:two-scenarios}
\end{figure}

In Figure \ref{fig:two-scenarios}, we demonstrated two common scenarios in our previous experiments.
A test scenario illustrated in Fig. \ref{fig:two-scenarios}(a) starts for the ego vehicle in the right-most lane.
The ego vehicle planned to change lane into the middle lane, and stay in that lane.
The other agent vehicle in front of it started in the middle lane and then starts changing lane into the right-most lane.
The vehicle control in the ego vehicle detects the front vehicle and tries to reduce its speed but not its direction.
The RSS requirements demands for safe steering behavior which in this case was not respected, and therefore this is a useful test.
In another situation demonstrated in Fig. \ref{fig:two-scenarios}(b), the ego initially is very close to the blue vehicle in front of it, and then the blue vehicle turns in front of it unsafely and collides with it.
This scenario is initially unsafe, and the ego vehicle should respond according to either of the lateral or longitudinal safety requirements.
If the initial speed of the ego vehicle is not too high, then it can brake on-time and satisfying the RSS requirements.
Otherwise, the ego vehicle cannot react on-time and contributes to the collision, which qualifies it to be collected as a useful test.
We emphasize that this class of test case results, which are hard for a human to detect, were one of the motivations for us to propose this approach for test generation.

\begin{table}
\begin{center}
\caption{{Statistical result about robustness of monitoring RSS and CAS specifications. We used ``pos'' as an abbreviation of positive and ``neg'' as an abbreviation for negative.
}}\label{tbl:comp-uds}
\resizebox{.35\textwidth}{!}{
\begin{tabular}{ |c|c|c|c| } 
\hline
\textbf{description} & \textbf{values} & \textbf{description} & \textbf{values}\\
\hline
Total pos (RSS) & 771 & Total pos (CAS) & 406\\
Total neg (RSS) & 229 & Total neg (CAS) & 594\\
Total zeros (RSS) & 0 & Total zeros (CAS) & 0\\
\hline
\multicolumn{4}{c}{\textbf{one-by-one comparison between RSS and CAS results }}\\
\hline
pos (RSS) pos (CAS) & 387 & neg (RSS) neg (CAS) & 210\\
neg (RSS) pos (CAS) & 19 & pos (RSS) neg (CAS) & 384\\
\hline
\end{tabular}
}
\end{center}
\end{table}

\begin{table}
\begin{center}
\caption{{The break down statistics of predicates that caused the highest negative robustness. 
}}\label{tbl:pred-stats}
\resizebox{.4\textwidth}{!}{
\begin{tabular}{ |c|c|c| } 
\hline
\textbf{predicates} &  \textbf{\# of violations} &  \textbf{description}\\
\hline
$S_{b,f}^{lon}$ & 2 & safe longitudinal distance \\[0.1cm]
$S_{l,r}^{lat}$ & 23 & safe lateral distance \\[0.1cm]
$A_{l,maxAcc}^{lat}$ & 67 & maximum allowed lateral acceleration \\[0.1cm]
$A_{l,minBr}^{lat}$ & 0 & minimum required lateral brake \\[0.1cm]
$V_{l,stop}^{lat}$ & 68 & zero $\mu-$lateral velocity \\ [0.1cm]
$V_{l,neg}^{lat}$ & 0 &  non-positive $\mu-$lateral velocity \\[0.1cm]
$A_{b,maxAcc}^{lon}$ & 33 & maximum allowed longitudinal acceleration \\[0.1cm]
$A_{b,minBr}^{lon}$ & 36 & minimum required longitudinal brake \\[0.1cm]
\hline
\multicolumn{3}{c}{\textbf{Execution Statistics}}\\
\hline
\scriptsize{violation \%} & $22.9\%$ & falsified percentage using the RSS rules \\
\hline
\end{tabular}
}
\end{center}
\end{table}

\begin{table}
\begin{center}
\caption{{Statistical result about robustness of monitoring RSS and CAS specifications in mostly unsafe situations. We used ``pos'' as an abbreviation of positive (satisfying specifications) and ``neg'' as an abbreviation for negative (violating specifications).
}}\label{tbl:comp-uacs}
\resizebox{.35\textwidth}{!}{
\begin{tabular}{ |c|c|c|c| } 
\hline
\textbf{description} & \textbf{values} & \textbf{description} & \textbf{values}\\
\hline
Total pos (RSS) & 406 & Total pos (CAS) & 20\\
Total neg (RSS) & 594 & Total neg (CAS) & 980\\
Total zeros (RSS) & 0 & Total zeros (CAS) & 0\\
\hline
\multicolumn{4}{c}{\textbf{one-by-one comparison between RSS and CAS results }}\\
\hline
pos (RSS) pos (CAS) & 0 & neg (RSS) neg (CAS) & 574\\
neg (RSS) pos (CAS) & 20 & pos (RSS) neg (CAS) & 406\\
\hline
\end{tabular}
}
\end{center}
\end{table}

In the next section, we discuss and present our experiments and their result.

\subsection{The necessity of RSS in testing}
For all the experiments in this section, we used the values in Table \ref{tbl:rss-params} and Table \ref{tbl:params}.
Also, the length of each simulated driving scenario is 10 seconds.
In our first experiment, we generated and stored one thousand uniformly random driving scenarios.
We do not precisely follow the executing sequences shown in Fig. \ref{fig:framework} as we are going to compare the result of monitoring the same test data against two different STL formulas.
Next, we computed the robustness of the test data against CAS specifications.

The result is demonstrated in the third and fourth columns of the upper part of Table \ref{tbl:comp-uds}.
About $40\%$ of the result did not violate their specification, and the other $60\%$ did violate their specifications.
The result of computed robustness of the test data against RSS specifications is shown in the first and second columns of the upper part of the same table.
The $23\%$ violated scenarios for the RSS case is noticeably less than the $60\%$ for the CAS case. 
The lower part of Table \ref{tbl:comp-uds} represents the statistics of all combinations of positive and negative results of each specification for each test.
Note that the total number of tests for which both specifications falsified and satisfied them is 210 and 387, respectively.
The result suggests that 384 falsified tests by the CAS specifications are not of our interest.
Furthermore, 19 tests were falsified with the RSS requirements, but not with the CAS requirements.
This implies that the test scenarios were potentially dangerous and hence worth further analysis, but they did not necessarily lead to a collision.
In this experiment, using the RSS specifications, we can disregard 384 meaningless and add 19 meaningful tests (about $40\%$ of the tests) that were misclassified by using a simple collision avoidance requirement (CAS).
Table \ref{tbl:pred-stats} denotes the list of predicates in the RSS formula and the number of times that each one got falsified with the highest negative magnitude (most violating).
A clustering algorithm can use the information about the falsified predicates to more tailor the test scenarios.

\subsection{Improving search-based testing through RSS}
In another experiment, we followed exactly the flow in Fig. \ref{fig:framework}.
We engaged the optimizer and used the CAS specification as its cost function (s.t. optimality means lower robustness for an input signal) to find more dangerous test driving scenarios.
We let the optimizer to find the worst falsifying behavior for the CAS specifications for one thousand iterations and stored the samples and their robustness result.
The result is presented in the third and fourth columns of the upper part of Table \ref{tbl:comp-uacs}.
Exactly $2\%$ of the result have positive robustness values, and the other $98\%$ have negative values.
The high percentage of negative result is because of the optimization.
The result of computed robustness of the stored test data against RSS specifications is shown in the first and second columns of the upper part of the same table.
The $60\%$ negative robustness for the RSS case is substantially less than the $98\%$ for the CAS case. 
The lower part of Table \ref{tbl:comp-uacs} represents the statistics of all combinations of positive and negative results of each specification for each test.
Note that the total number of tests for which both specifications falsified and satisfied them is 574 and 0, respectively.
The result suggests that 406 falsified tests by the CAS specifications are not of our interest.
Furthermore, 20 tests were falsified with the RSS requirements but not with the CAS requirements.
In this experiment, using the RSS specifications, we can disregard 406 meaningless and add 20 meaningful tests (about $42\%$ of the tests) that were misclassified by using a simple collision avoidance requirement.

\subsection{Falsification by RSS}
In the last experiment, according to the diagram in Fig. \ref{fig:framework} we engaged the optimizer and used the RSS specification as its cost function to find more relevant test driving scenarios.
We let the optimizer to try to optimally falsify the RSS specifications for 350 iterations and stored the samples and their robustness result.
Similar to what we did in the last two experiments, we repeated the process for the CAS specifications.
The result are depicted in Table \ref{tbl:comp-opt-rss} and Table \ref{tbl:pred-opt-stats}.
The optimizer is especially helpful when falsification is the priority.
An alternative approach could be running a script that screens all the randomly generated tests and computes and classifies their safeties.
However, this approach can take longer to falsify the specifications in comparison to using an optimizer.

The predicate statistics in Table \ref{tbl:pred-stats} and Table \ref{tbl:pred-opt-stats} suggests that we can further classify test scenarios  based on their violated constraints.
For example, we might only be interested in unsafe test scenarios in which the maximum allowed lateral acceleration denoted by predicate $a_{maxAcc}^{lat}$ was violated by the controller.

\begin{table}
\begin{center}
\caption{{Statistical result about robustness of monitoring RSS and CAS specifications where we used the RSS rules for optimization. We used ``pos'' as an abbreviation of positive and ``neg'' as an abbreviation for negative.
}}\label{tbl:comp-opt-rss}
\resizebox{.35\textwidth}{!}{
\begin{tabular}{ |c|c|c|c| } 
\hline
\textbf{description} & \textbf{values} & \textbf{description} & \textbf{values}\\
\hline
Total pos (RSS) & 292 & Total pos (CAS) & 50\\
Total neg (RSS) & 58 & Total neg (CAS) & 300\\
Total zeros (RSS) & 0 & Total zeros (CAS) & 0\\
\hline
\multicolumn{4}{c}{\textbf{one-by-one comparison between RSS and CAS results }}\\
\hline
pos (RSS) pos (CAS) & 49 & neg (RSS) neg (CAS) & 57\\
neg (RSS) pos (CAS) & 1 & pos (RSS) neg (CAS) & 243\\
\hline
\end{tabular}
}
\end{center}
\end{table}

\begin{table}
\begin{center}
\caption{{The break down statistics of predicates that caused the highest negative robustness in the last experiment. 
}}\label{tbl:pred-opt-stats}
\resizebox{.4\textwidth}{!}{
\begin{tabular}{ |c|c|c| } 
\hline
\textbf{predicates} &  \textbf{\# of violations} &  \textbf{description}\\
\hline
$S_{b,f}^{lon}$ & 2 & safe longitudinal distance \\[0.1cm]
$S_{l,r}^{lat}$ & 3 & safe lateral distance \\[0.1cm]
$A_{l,maxAcc}^{lat}$ & 13 & maximum allowed lateral acceleration \\[0.1cm]
$A_{l,minBr}^{lat}$ & 0 & minimum required lateral brake \\[0.1cm]
$V_{l,stop}^{lat}$ & 0 & zero $\mu-$lateral velocity \\ [0.1cm]
$V_{l,neg}^{lat}$ & 0 &  non-positive $\mu-$lateral velocity \\[0.1cm]
$A_{b,maxAcc}^{lon}$ & 35 & maximum allowed longitudinal acceleration \\[0.1cm]
$A_{b,minBr}^{lon}$ & 5 & minimum required longitudinal brake \\[0.1cm]
\hline
\multicolumn{3}{c}{\textbf{Execution Statistics}}\\
\hline
\scriptsize{violation \%} & $16.5\%$ & falsified percentage using the RSS rules \\
\hline
\end{tabular}
}
\end{center}
\end{table}

%% file: conclusion.tex
\section{Conclusion}

In this paper, we present an automated and qualification based method for generating driving test scenarios.
The generated tests could be used for discovering control software bugs in autonomous vehicles. 

We integrated the RSS requirements into our Sim-ATAV framework, which is publicly available as part of \staliro tool.
Now it is possible to test the control and perception system stack against the RSS model.